**Lessons in Co-Creation: the Inconvenient Truths of Inclusive Sign Language Technology Development**

Maartje De Meulder[1], Davy Van Landuyt[2], and Rehana Omardeen[3]

In the era of AI-driven language technologies, the participation of deaf communities in sign language technology development – often framed as "co-creation" – is increasingly emphasized. We present a reflexive case study of two Horizon 2020 projects on sign language machine translation (2021-2023), conducted with a EUD, a European-level deaf-led EUD. Using participant observation, internal documentation, and collaborative analysis among the authors, we interrogate co-creation as both a practice and a discourse. We offer five lessons for making co-creation consequential: (1) recognize and resource deaf partners' invisible labor; (2) manage expectations via accessible science communication; (3) "crip" co-creation by dismantling structural ableism; (4) diversify participatory methods to address co-creation fatigue and intersectionality; and (5) redistribute power through deaf leadership. We contribute an empirically grounded account of how co-creation plays out in multi-partner AI projects, and actionable implications for design that extend to participatory AI with minoritized language and disability communities.

## 1 INTRODUCTION

In the era of AI-driven language technologies, there is a growing call for the participation and leadership of deaf communities in the development of sign language technologies. This imperative is frequently framed in terms of "co-creation", a term used across research and industry to describe processes involving end-users and stakeholders throughout a project's lifecycle. In response to a 2018 EU Horizon 2020 funding call focused on mobile translation applications between signed and spoken languages (ICT-57-2020), the European Union of the Deaf (EUD), a European deaf-led NGO participated in two large-scale projects: EASIER and SignON. Spanning from 2021 to 2023, these projects aimed to develop sign language machine translation (SLMT) systems through multi-partner consortia comprising predominantly hearing, non-signing researchers, and a smaller number of deaf stakeholders. The Horizon's program explicit requirement for user validation catalyzed the EUD's involvement as an essential stakeholder, which enabled the organization to secure funding as a consortium member, and shape research priorities [39].

---

[1] HU University of Applied Sciences, the Netherlands
[2] European Union of the Deaf
[3] European Union of the Deaf

While co-creation is often presented as an ideal or even required methodology for inclusivity and responsible innovation, the actual experiences of deaf partners in such projects reveal a complex picture. This complexity is acknowledged to some extent by at least SignON, which issued a white paper on the do's and don'ts of "inclusive collaboration among policy makers, researchers and end-users" in the context of sign language technologies [45]. Our paper unpacks this further and presents an in-depth reflection on the involvement of the EUD in EASIER and SignON, based on informal participant observations, internal documentation, and collaborative discussions and analysis among the authors. We interrogate co-creation as both a method and discourse, uncovering several 'inconvenient truths' about the challenges of applying co-creation in practice. In doing so, we contribute to ongoing discussions about ethical and effective design of SLMT systems [25,33,53,54]. This paper offers five lessons for future co-creation initiatives in this space and argues for reimagining co-creation as a transformative practice that shifts power relationships and fosters deaf leadership in sign language technologies research and development.

We structure the paper as follows: section 2 outlines related work from sign language technologies, participatory design, and disability justice. Section 3 details our methodological approach. Sections 4 and 5 present challenges during co-creation and for deaf partners respectively. Section 6 synthesizes these into five actionable lessons, and section 7 concludes ty reframing co-creation as a transformative, rather than participatory, endeavor.

## 2 RELATED WORK

Co-design and co-creation are widely used to involve stakeholders in technology development, typically emphasizing collaborative efforts across a project's lifecycle. While the terms are often used interchangeably [51], co-design (or participatory design) tends to focus on the collaborative design phase using workshops, prototypes, and mock-ups, whereas co-creation extends across the entire lifecycle of a project, employing broader iterative methods such as brainstorming sessions, focus groups, testing, and iterative prototyping [11,31]. Both approaches aim to integrate stakeholders' lived experiences in technology development.

In sign language technology research, co-creation or co-design is mostly conceived of what Sloane et al. call "participation as consultation" [47]. It typically involves iterative processes such as comprehension testing through surveys [24,43], focus groups, interviews, round tables, iterative co-design sessions, live co-creation events [14,21,26,39], and evaluations through questionnaires, semi-structured interviews, TAP protocols, users observation, or Wizard of Oz techniques [42]. Typically, deaf users are positioned here as testers of pre-built systems rather than as co-designers shaping requirement from the earlies stages of the AI lifecycle [30].

The field of SLMT as well, has witnessed a "participatory turn" [19], increasingly emphasizing participatory methods [9,15,25,45,55]. Yet it remains dominated by hearing, often non-signing, researchers and developers. Co-creation is frequently named as ethical or emancipatory, but critical work points to significant power imbalances and extractive practices [8,12,19,22,33,35,46]. Studies have shown that while deaf participants are included, their roles rarely involve leadership or decision-making authority [4], amounting at times to "participation washing" [47] or "pseudo-participation" [37]. This superficial inclusion contributes to epistemic injustice and reinforces systemic inequities [3,20].

These critiques resonate with broader debates in HCI, where participatory design has long been scrutinized for tokenism and designer-expert hierarchies. Critical perspectives such as crip technoscience [28,29], design justice [13], and Disability Justice [7] highlight the need to center disabled people as knowledge producers and leaders. Scholars have argued for situated, relational approaches that dismantle structural ableism and redistribute power in design processes [50]. Deaf scholars similarly call for deaf-led research agendas in sign language AI [15,16,20], directly addressing epistemic justice and structural inequities in the field. Together, this body of work emphasizes that co-creation should not be reduced to mere participation, but as a reconfiguration of who is recognized as a designer and decision-maker

## 3 METHODS

### 3.1 Research approach

We adopt a collaborative reflexive case methodology, aligned with interpretivist traditions in HCI [48]. Our aim is not to test hypotheses but to generate situated, actionable understanding of how co-creation happened, in a specific moment in time, in two large mixed deaf/hearing, multi-partner consortia. The two projects used as case studies, EASIER and SignON, provide a significant case for studying co-creation in sign language AI research due to their size, scope, and centrality in the European research landscape during the funding period. Analytically, we moved back and forth between what we observed during the projects, concepts from design justice, Deaf Studies, and crip technoscience, and the perspectives we had at the time of our discussions. While the projects took place between 2021-2023, most of our collaborative discussions occurred in 2023 and 2024. This means we were not only analyzing the projects as they unfolded, but also retrospectively, with the additional knowledge and experience we had acquired since. This iterative process allowed us to identify recurring patterns and make sense of them in relation to broader theories. Rather than starting with fixed categories, we allowed insights to emerge through this dialogue between our data and existing frameworks. In doing so, we generated mid-level concepts that inform both design practice and governance.

### 3.2 Data sources

Data for this study were collected through

(a) Vignetted and reflections. We shared stories and reflections with each other, both signed (mostly in International Sign and a specific national signed language), and written (mostly through

email, in English). These came up during conference side-meetings, work package meetings, informal conversations, and several dedicated sessions for this paper. We did not produce verbatim transcripts of the signed exchanges but summarized them as analytic memos after conversations.

(b) Participant observations gathered mostly by Davy (deaf) and Rehana (hearing) both staff members at the EUD during the project durations. Discussion with Maartje, a deaf academic, further contextualized and deepened the data.

(c) Project artifacts: we consulted deliverables, reports, slide decks, meeting notes as additional context to triangulate our analysis.

Together, these sources provide complimentary perspectives on how co-creation was organized and experienced across both consortia.

### 3.3 Collaborative analysis

Over the course of the projects we discussed our observations with each other informally, which gradually developed into the idea for this paper. As the projects neared their end, the collaborative process intensified. Our analysis was dialogic and reflexive: we compared perspectives, challenged each other's assumptions, and drew on different standpoints and lived experiences (deaf/hearing, advocacy/academic, insider/outsider). Rather than coding systematically, we identified patterns that explained recurring experiences across projects. These patterns were consolidated into five lessons for designing co-creation processes. We also presented preliminary findings at an international conference in 2023, which helped refine our framing.

### 3.4 Ethics and positionality

This research reflects on organizational practices in two EU-funded consortia involving small, tightly connected deaf and hearing networks. The second and third authors worked as staff at EUD during EASIER and SignON and remain employed by the same NGO; the first author is a deaf academic collaborator outside the EUD. These roles afforded privileged access to co-creation processes and artifacts but also introduce potential institutional-loyalty and advocacy biases. To address this, we worked reflexively as a team: questioning each other's assumptions, sharing standpoints, and practicing "critical friend" analysis across our different positions. This dialogic process helped us balance insider perspectives with external critique. The views expressed here are those of the authors alone, and not official positions of consortia or funders. Funding bodies had no role in the study design, analysis, or writing of this paper.

### 3.5 Limitations

We acknowledge that our data are limited to two (admittedly large) European SLMT projects, and to the experiences of a relatively small number of staff and contributors. This inevitably limits the scope of our findings. For instance, we could not conduct detailed conversational analyses, and our accounts are shaped by the perspectives of those directly involved. Some of our reflections also rely on recall, which introduces potential bias. Our findings should thus not be read as representing all deaf or hearing participants in these projects, nor the full diversity of experiences in SLMT.

Our aim is to offer grounded, situated lessons that may be useful beyond these two cases. Readers should judge for themselves how our analysis may transfer to other participatory AI projects involving disability and/or minoritized language communities.

## 4 EMPIRICAL CONTEXT: STRUCTURING THE FIELD OF SIGN LANGUAGE AI

In 2018, the European Union launched a Horizon 2020 funding call specifically targeting the development of translation applications between signed and spoken languages. Two large-scale projects, EASIER and SignON, were funded under this call, rapidly concentrating much of the European SLMT community by channeling funding, technical expertise, and organizational influence into their consortia. Although distinct projects, they were interconnected through shared events and policy discussions, reinforcing their collective impact on research priorities, methods, and ethical narratives in SLMT.
Each consortium was large – 14 organizations for EASIER and 17 for SignON, with over 60 individual members in each consortium – and brought together technology experts (predominantly hearing, non-signing) with sign language academics (predominantly hearing and signing), and organizations representing deaf end-users (predominantly deaf signers). The size and structure of these partnerships make them significant cases for examining structural patterns in sign language AI research in Europe. During their three-year duration (2021-2023), most researchers in the field in Europe were involved in one of these two projects, reflecting the novelty and concentration of the field. This convergence of expertise offered a unique opportunity to capture a snapshot of the field at a specific moment in time.

EASIER and SignON were the first large-scale European initiatives to tackle sign language machine translation across several spoken and signed languages in Europe. Their proposals, developed around 2020 (and thus reflecting the state-of-the-art at that time), set out ambitious goals that stretched the technological limits then available. By the time the projects concluded in 2023 – and with hindsight from 2025 – it is clear that these ambitions were both visionary and, in some respects, premature. This temporal context matters for our analysis: it highlights how quickly technological capacities and community expectations have evolved in just a few years, and why the co-creation practices employed during these projects took the form they did. These projects demonstrate how co-creation is shaped not only by community involvement by also by prevailing technological imaginaries, which set the boundaries of what was considered possible or desirable at the time.

The deaf-led EUD is a long-established advocacy organization and was the only deaf-led partner involved in *both* consortia. While highly experienced in European policy and rights-based advocacy, the EUD entered these projects with limited prior technical expertise in AI or machine translation. Their involvement was driven by a recognition of the importance of influencing how these emerging technologies might impact deaf communities, alongside the opportunity to build internal expertise in engaging with sign language technology development.

However, participating in large, hearing-dominated, highly technical, multi-partner consortia posed challenges. These included navigating technical language and decision-making structures, balancing advocacy with operational project responsibilities, and managing expectations about what meaningful participation could realistically look like in such settings. This structural context directly shaped the dynamics of co-creation that unfolded through the projects. It is within this field-shaping, complex environment that the following examines how co-creation operated in practice, and what tensions and frictions emerged for partners like the EUD. We will now briefly discuss EASIER and SignON, and the EUD's role in them.

### 4.1 EASIER, SignON, and the EUD's role

EASIER was a three-year project funded by the EU Horizon 2020 program with a grant of €4 million [41]. It focused on developing end-to-end translation for seven European signed languages and their corresponding spoken languages, envisioning use that centered on supporting deaf and hearing users in the workplace and everyday life. The project targeted different technology readiness levels (TRLs) for different components of the translation framework, ranging from TLR7 (system prototype) to TLR8 (complete system).

SignON was also a three-year project funded by the EU Horizon 2020 program, funded with €5.6 million over the same period. SignON aimed to develop a mobile application for automatic translation between various European signed and spoken languages, with use cases focused on low-risk, low-stake contexts, with the main demonstration case being the hospitality industry. Its goal was to reach TLR7; by project completion, TLR6 was achieved (insert ref SignON consortium 2023a). One of the distinct features of SignON was its (self-reported) "strong commitment to the co-creation principle", with co-creation defined as "a collaboration between researchers, developers, and users, based on continuous, periodic information exchange, expectation management, openness and user-involvement (in the design and development process), on equal merits and built on trust, from the project inception" [44].

Critically, neither EASIER nor SignON featured a deaf academic as one of the Principal Investigators.

The timeline and degree of the EUD's involvement vary across both projects. For SignON, a meeting of convenience was organized around an existing academic conference in Brussels, six months before the call's deadline. For EASIER, the EUD's involvement started only a few months before the deadline. In both SignON and EASIER, the EUD was responsible for leading the respective WP1 focusing on the co-creation aspect of the projects. Additionally, they shared feedback and expertise on areas directly related to their expertise such as accessibility, sign languages, and the impact of sign language technologies on deaf communities. In the EUD's experience, attending preparatory meetings, getting your foot in the door early, and having spontaneous individual calls with potential partners considerably increased the chances of exerting influence on the work plan, and securing a leading role with the consortia. Indeed, "I had a quick call with …" is one of the most common phrases heard during consortium formation – yet these "quick calls" often happen over the head of deaf partners. While various partners in EASIER and SignON had deaf staff members, especially those with a longer tradition in sign language research, the EUD and a sign language center in one of the partner countries (name withheld for review) were the only organizations with deaf staff in leadership roles. The sign language center was primarily responsible for communication and dissemination within SignON. Individual deaf contributors to EASIER and SignON had various roles, including language and communication experts, (unpaid) advisory board members, and co-creation facilitators, mainly consisting of using their personal networks to recruit local deaf staff members to facilitate evaluations.

Within EASIER, the EUD led the work package "User involvement". For this, collaborating with other partners, they conducted initial focus groups with both deaf and hearing end-users at the beginning of the project, and planned two further cycles of end-user evaluations across the different language communities, to gather user feedback on the technology's performance [38,40]. The first evaluation cycle, halfway through the second project year, consisted of focus group discussions led by a facilitator. Participants looked at examples of the technology and then discussed them. They did this for the mobile app prototype, the avatar, and the output of the early gloss-based translation models. The second evaluation cycle, halfway through the third project year, was a more structured evaluation focusing on the themes identified in the first evaluation. For the mobile app and the avatar, participants completed a survey and were also able to give feedback in a discussion format. Translation models were evaluated separately by sign language translation professionals. Three components of the EASIER technology were independently tested in both evaluations: the machine translation models, the avatar, and the mobile application (for usability and design). The evaluation usually followed a two-step format, combining a rating task and a focus group discussion. A focus group discussion was used to collect qualitative feedback, aiding in design improvements, such a refining mouthing of the avatar and aspects related to user interface (click flow of the app). Quantitative methodologies were used to measure the achievement of predefined project key performance indicators. For different technologies different quantitative methods were used: the mobile application was rated using a System Usability Scale; the avatar via a Likert rating task, and translation models using a rating task common to MT research. These evaluations took place in each of the target countries with both deaf and hearing participants. Evaluations were led by local facilitators from the partner institutions who recruited participants from the local signing community. Participants were paid for taking part in the evaluation.

Within SignON, the EUD led the WP "co-creation and user response". In this capacity, they collaborated closely with other partners to collect user feedback using various methods including focus groups, online surveys, and live co-creation events across the different sign language communities targeted by the project. The aim of these focus groups was twofold: (1) to identify specific use cases or circumstances in which machine translation apps would or would not be used, and (2) identify the needs, expectations and experiences of deaf participants regarding translation technology. The aims were like those of EASIER's initial focus groups, with the only difference being that EASIER also targeted sign language professionals, while SignON did not. In addition to these initial focus groups, the EUD also organized separate focus groups and interviews to better understand the needs of minorities within deaf communities, such as deafblind people and ethnic minorities, in relation to sign language technologies. For this data collection, similar questions were asked as for the more general focus groups, however interviewees were aware that the aim was to collect perspectives from people of specific intersecting identities. The EUD also organized co-creation events in collaboration with local deaf organizations and consortium partners. These events were open to the public although most were relatively small (max. 30 people including project staff). During these events, the EUD and consortium partners opened with presentations about the project and the technology, followed by questions and discussion with the audience. Audience members were not paid for participation; however, they did sign consent forms (provided in local written and sign languages) to allow for their input to be used by the project. These events served as platforms to engage deaf end-users, providing them with project-related information and context, while also gathering their ideas, feedback and concerns about the technology being developed. By involving local technology partners to directly present to deaf audiences, these events also fostered direct engagement between developers and end-users.

## 5 CHALLENGES DURING THE CO-CREATION PROCESS

In this section, we analyze the challenges that emerged during the co-creation processes of EASIER and SignON, particularly highlighting the tensions and labor experienced by deaf partners.

### 5.1 Managing expertise gaps and collecting meaningful feedback

One fundamental challenge in co-creation is collecting meaningful feedback on technologies still under development. For many lay participants, sign language technology remained surrounded by misconceptions. For instance, the term "avatar" was frequently used to refer to the entire SLMT system, conflating the visual output with the underlying translation processes. Participants often lacked a clear understanding of distinctions between mocap-generated avatars and rule-based avatars. This limited understanding sometimes complicated feedback collection, as participants tended to focus on superficial or irrelevant aspects of prototype demonstrations.

This highlighted the need for robust science communication during co-creation, providing appropriate background information tailored to participants' prior knowledge. While lay participants required more explanatory context to offer informed feedback, participants with technical expertise were better able to provide targeted feedback. Structured and well-framed feedback instruments – such as rating scales– were essential not only for guiding user input but also for benchmarking technological progress across different iterations. Critically, even negative ratings served a purpose: they offered concrete evidence to funders about the immaturity of the technology and the importance of investing in and directing resources to foundational research, developing component parts of lower TRLs and intensifying sign language data collection efforts before advancing to full translation systems.

### 5.2 Redundancy, overlap, and co-creation fatigue

With both projects running simultaneously and often targeting similar deaf communities across Europe, participants were repeatedly asked to provide feedback on early-stage systems. This redundancy created concerns among the consortium partners about resource use and participant fatigue, particularly when early prototypes offered limited functionality and feedback sessions yielded similar outcomes across contexts. Consequently, co-creation could be perceived as performative in nature. Nevertheless, these repetitive engagements also served to introduce communities to emerging technologies and foster general awareness. Linguistic variation across communities (e.g. gender- or identity-related signs) added layers of complexity, occasionally resulting in valuable localized insights, despite the perceived repetition [32].

### 5.3 Cautious co-creation and risk management

Some partners approached community engagement with caution, shaped by prior negative experiences. For example, one previous project was poorly received and met with resistance received because it showcased avatar animations that were essentially just moving stick figures, known as pose videos. Partners involved in that project were resistant to present similar pose videos to users in a MT rating task. This caution manifested as hesitancy to share work-in-progress prototypes for fear of damaging community trust. In response, the EUD often played a gatekeeping role, advising on what to present and how to frame it. Sometimes, this elevated the EUD's voice even above other deaf participants in decision-making discussions. However, the EUD also faced limits in representing all sub-groups within deaf communities. For example, there were discussions about physical representations of a non-binary sign language avatar that the EUD staff felt ill-positioned to answer on behalf of any non-binary deaf individuals. Attempts were made to include underrepresented voices through targeted recruitment, though these efforts were sometimes constrained by limited timelines and resources.

Notably, the two projects differed structurally: EASIER included built-in checkpoints requiring iterative community feedback, while SignON lacked systemic mechanisms for integrating end-user feedback during technical development, leading to co-creation events that remained largely conceptual without generating concrete feedback.

### 5.4 Anonymity and dense deaf networks

Giving feedback on sign language technologies can be taxing for participants, since they may relate the use of those technologies to their own lived experiences, personal stories, and even traumas. Ensuring and maintaining anonymity in focus group and interview data is thus critical, but also presents challenges, particularly given the dense, interconnected nature of many European deaf communities. Even with personal identifiers removed, details such as prior project involvement, nationality, country of residence, or gender identity could easily reveal participants' identities. These risks were particularly salient when recruiting participants with specialized expertise (e.g. translation professionals or deaf technologists). Safeguarding anonymity required additional measures beyond standard anonymization protocols, including careful vetting of participant combinations and cross-referencing potentially identifiable traits. This demonstrates how ethical commitments to participant safety intersect with the practicalities of recruitment and facilitation.

### 5.5 Recruitment and role of local deaf facilitators

Local deaf facilitators were crucial for successful participant recruitment, leveraging personal and professional networks within national communities. Intersectional identities of facilitators also shaped recruitment outcomes: for example, non-binary facilitators successfully recruited non-binary participants. Strategic partnerships with community organizations, such as (name of organization withheld for review), further supported inclusive recruitment of ethnic minority participants. These relational dynamics were indispensable for assembling diverse, representative feedback groups, but also made recruitment inseparable from broader questions of trust, identity, and anonymity.

### 5.6 Invisible labor and unequal access responsibilities for deaf partners

Deaf partners – especially the EUD – carried a disproportionate share of invisible labor throughout the projects. This included supporting less experienced hearing partners in learning how to engage with deaf communities, tailoring communication materials for deaf audiences, and drawing on personal networks to fill recruitment gaps. For example when it appeared no Spanish SignON partners had connections with the Spanish deaf community, AUTHOR, originally from (country withheld for review) but now living in (country withheld for review), used their personal network to help organize the co-creation event in (city withheld for review).

Additionally, organizing linguistic accessibility for meetings required substantial coordination: interpreter scheduling, working across multiple sign languages, securing qualified interpreters for highly technical discussions, and accommodating last-minute changes were regular demands. Partners with less experience often overlooked the need for sign language interpretation in online and in-person meetings, requiring deaf partners to arrange this themselves. Even experienced partners had to adapt their working practices and routines. This included for example working with International Sign interpreters, who may need longer booking times and work within a different fee system. Even with interpreters present, deaf participants often had to adapt their signing to match interpreters' repertoires and navigate varying levels of interpreters' familiarity with SLMT content. Given the size of the consortia, a significant amount of information exchange occurred through email, especially when working under the time pressure of tight deadlines. Last-minute meetings were often particularly difficult to arrange, due to key consortium members relying on interpreting services (and thus interpreter scheduling) and the alignment of key members' schedules.

This extensive but often undocumented labor placed considerable strain on deaf partners, spreading already limited personnel thinly across multiple responsibilities.

### 5.7 Financial and organizational burden for accessibility

Co-creation processes involving deaf and hearing partners present challenges, particularly when partners have varying levels of signing skills and varying levels of experience working with deaf communities and signed languages. Financial responsibility for linguistic accessibility frequently became an issue. Responsibility for funding accessibility varied between projects. In EASIER, the EUD managed its own accessibility budget and coordinated interpreter services, while in SignON, host organizations were expected to cover accessibility costs, occasionally resulting in logistical tensions when different national sign languages were involved. It was crucial to distinguish between internal accessibility costs (e.g. interpreters at meetings) and external costs (e.g. translated research materials or communication campaigns), especially when working with different signed languages. This distinction could sometimes become a sticking point. Some partners initially attempted to shift financial responsibility for local translation or accessibility needs onto the EUD, though most partners ultimately managed these obligations independently. Nonetheless, the variability of access arrangements introduced ongoing deaf tax for deaf partners, who had to advocate for consistent, adequate accessibility throughout the projects.

### 5.8 Broader structural asymmetries in deaf-hearing consortia

Underpinning these challenges were larger structural asymmetries characteristic of hearing-dominated research consortia. The absence of deaf PIs in both projects reflects deeper systemic barriers to deaf leadership in sign language technology development in Europe. These asymmetries shaped not only co-creation processes but also career pathways, labor expectations, and epistemic authority through the projects.

All of this highlights that deaf partners contributed more than what is explicitly documented. Some of this labor was largely invisible. This included for example providing critical insights into cultural differences, supporting hearing partners' learning curves, educating others about accessibility and effective recruitment strategies for deaf participants, and organizing access. Other labor was more visible, for example using personal connections to recruit participants. As such, deaf partners performed additional (visible and invisible) labor that hearing partners were not required to do or didn't even think of. The lack of a critical number of deaf partners meant that those involved were at times spread too thin, impacting their ability to manage additional responsibilities. Consequently, the substantial time deaf partners devoted to these additional responsibilities at times hindered their progress on core duties. This diversion of (paid) working time to unplanned tasks could adversely impact the quality of output. In some cases, the EUD simply did not have the bandwidth to engage in coaching partners to improve their approach to deaf communities, given all the other invisible labor already required of them.

## 6 FIVE LESSONS FOR TRANSFORMATIVE CO-CREATION

The inclusion of deaf-led organizations such as the EUD in SLMT research projects has led to significant positive changes both within and outside the consortia. Within the consortia, this involvement has facilitated cross-disciplinary exchanges crucial for addressing technical and ethical challenges, such as ensuring anonymity in video recordings. The participation of deaf-led organizations has also fostered more positive and receptive attitudes towards these projects among deaf end-users, who felt their needs and perspectives were being genuinely represented. This co-creation approach has enabled the EUD to formulate several useful recommendations for developing SLMT technologies, centering deaf end-user experiences in the development process, and fostering meaningful discourse between technology experts and organizations representing deaf communities. From these experiences, 'best practice' guidelines have emerged [45], providing a valuable resource for future projects in this domain. In July 2025 the EUD, following the organization of an international workshop on sign language AI, published a book on sign language AI, exploring its impact on deaf communities' digital rights [52]. This shows capacity-building and increase in AI literacy inside the EUD thanks to involvement in these projects.

However, there are a few inconvenient truths of co-creation processes that future SLMT projects will be confronted with and need to address. Below, we list the five lessons that might be learned from them. These lessons, which are interrelated, reflect both the practical and structural challenges of co-creation in sign language AI research.

### 6.1 Lesson 1: recognize and resource deaf partners' invisible labor

Co-creation frequently demands both visible and invisible labor from deaf partners at multiple levels. Within Deaf Studies, this labor has been conceptualized as "deaf tax" [1,27], reflecting the systemic, often unacknowledged effort required of deaf individuals to operate within hearing-centric structures. This phenomenon has further been substantiated by Angelini, Spiel, and De Meulder [4], who, through interviews with staff members in SLMT projects, identified an implicit expectation for those staff members to take on additional responsibilities. In some cases, this extended to epistemic exploitation [6], wherein the specialized knowledge and lived experience of deaf staff members are drawn upon without adequate recognition or compensation. Deaf partners provided critical insights into cultural differences, supported learning curves for hearing partners, educated others about accessibility, recruited participants through personal networks, and managed access logistics. These forms of labor – though essential – were rarely explicitly recognized or resourced.

Even when interpreters were provided, additional labor was required to coordinate, adapt signing styles to interpreter repertoires, and process technical content through multiple layers of interpretation. Hearing partners often underestimated these demands, perceiving access had been fully addressed through interpreter provision alone – an issue not exclusive to such consortia [18]. The cumulative effect of these invisible burdens sometimes spread deaf staff too thin, diverting time from core responsibilities and leading to burnout or withdrawal from projects. Acknowledging this labor requires more than rhetorical commitments to inclusion. Structural provisions must be built into project budgets, timelines, and role distributions to account for these additional responsibilities. A failure to acknowledge and accommodate these demands risks reinforcing structural inequities and undermining the sustainability of collaborative research efforts. Therefore, it is imperative that future projects proactively account for this labor – both in terms of budgeting and structural support. Even just making everyone aware that it is *work* constitutes a critical step towards equitable research collaborations.

Our intention here is not to make an accusation of deliberate exploitation, but to highlight systemic disparities in how labor is distributed in sign language AI projects. The issue is not simply that different partners have different roles, but that certain types of labor – particularly those requiring deeper engagement with deaf communities – are often disproportionately assigned to deaf partners, without adequate recognition or resourcing. While project management structures may allow for reporting these discrepancies, our findings suggest that these imbalances often go unnoticed or unchallenged because they are normalized within the field.

### 6.2 Lesson 2: manage expectations through transparent and accessible communication

Co-creation in emerging technologies like SLMT requires carefully managing expectations among all stakeholders. Funding applications in this domain often involve optimistic portrayal of technological capabilities, leading to heightened public expectations that are difficult to meet during early stages of development [49]. This creates a tension between hyped narratives and the realities of technological immaturity.

Participants unfamiliar with SLMT often conflate avatars with full translation systems, may misunderstand the complexity of underlying models, or expect immediate functionality. Clear science communication is therefore critical – not as a peripheral activity but as an integrated component of co-creation [5,36]. Explainable AI principles [2,23,34] must be applied not only to system design but to participatory processes: participants need honest, accessible explanations of what technologies can and cannot yet do. Failure to manage expectations risks generating false hope and exacerbating co-creation fatigue. Transparent communication is essential to ensure participants feel informed, respected, and aware of how their feedback contributes to incremental development.

### 6.3 Lesson 3: crip co-creation by dismantling structural ableism

The lessons learned from EASIER and SignON highlight broader patterns in AI and accessibility projects aimed to assist users with disabilities, where governance and decision-making are often concentrated among non-disabled Principal Investigators, with most consortium members reflecting this demographic. Despite the growing emphasis on co-creation in accessibility research, many projects continue to reinforce ableist assumptions in their design and implementation. "Cripping" co-creation requires dismantling these assumptions form the outset. This includes rethinking timelines that privilege last-minute decision-making, accommodating diverse working capacities, and challenging ingrained expectations about how research is designed, who gets to participate, and whose expertise is valued. As the EUD's experience demonstrates, exclusion from early-stage planning directly limits influence over funding distribution, work plans, and research priorities. Cripping co-creation is not only about access accommodations but about restructuring entire research cultures. In doing so, these changes stand to benefit all partners, not just those with disabilities.

### 6.4 Lesson 4: diversify methods to address co-creation fatigue and intersectionality

Co-creation in technology development exists within a framework where iterative testing and feedback loops are common and systems are constantly adapting to new training input, requiring repeated feedback from participants. Co-creation in these contexts raises critical questions about what can reasonably be expected of stakeholders [19,22]. EASIER and SignON concentrated significant funding and resources in a handful of relatively small European countries. The demand for co-creation will not decrease. With the proliferation of sign language AI projects, deaf communities increasingly face "co-creation fatigue" – repeatedly asked to contribute feedback without sufficient recognition, compensation, or immediate benefit. This is particularly relevant when co-creation involves addressing barriers that participants themselves experience daily (e.g. access or communication challenges), while project outcomes remain abstract, distant, or lacking immediate applicability. Reimagining participatory methods is essential to sustain meaningful engagement.

Innovative, non-traditional methods – such as art-science collaborations, speculative design, or performance-based engagements – can revitalize participation while reducing repetitive feedback fatigue and offer more diverse, sustainable, and creative approaches to co-creation. (Name withheld for review) [36] exemplifies how creative engagement can open richer conversations about technology's implications without relying solely on technical evaluation protocols.

Flexible methods also better support intersectional recruitment. For example, targeted partnerships enabled the inclusion of non-binary and ethnic minority deaf participants who might otherwise be underrepresented. Group dynamics varied depending on participants' age, technological literacy, and comfort levels, requiring adaptive facilitation. Universal design principles applied to co-creation protocols, such as the adapted "Think-In" format [10] demonstrate that accessibility-driven modifications often benefit all participants.

Effective co-creation requires not only inclusive recruitment but the flexibility to tailor methods to diverse participant needs, avoiding one-size-fits-all models of engagement. Expanding methodological repertoires supports both ethical participation and richer data collection. For instance, while group interviews may be preferred by sighted deaf participants, some deafblind participants found 1:1 interview settings more accessible. Similarly, older deaf participants sometimes felt more comfortable expressing themselves in 1:1 settings, rather than participating in discussions with younger people they perceive as more technologically savvy. These nuances highlight the necessity of adaptive co-creation strategies that acknowledge different accessibility needs, lived experiences, and power dynamics within deaf communities.

Additionally, co-creation processes must engage with deaf people from diverse backgrounds to ensure that technology is widely acceptable and usable. While all deaf people might, in theory, be end-users, their experiences, expectations, and technological literacy vary significantly. Co-creation processes must therefore consider who is involved – deaf experts or deaf lay people – as their understandings and reactions to technology can differ. For example, differences in understanding why certain technological issues exist – such as the shortage of quality sign language data – can shape varied feedback and expectations. Simply engaging "deaf participants" as a homogenous group is insufficient; intersectional perspectives must be actively integrated into co-creation efforts.

### 6.5 Lesson 5: shift to deaf leadership and structural redistribution

Co-creation projects can sometimes fall into the trap of virtue signaling, where the appearance of inclusivity and collaboration masks systemic issues of power imbalances and tokenism. Many sign language technology projects are designed and led by hearing researchers, who may not always fully understand or prioritize deaf communities' perspectives. This dynamic shapes decision-making processes and research outcomes in ways that reflect hearing researchers' biases and institutional priorities rather than the needs and interests of deaf partners [20]. Superficial inclusion is one of the most common manifestations of this problem. Deaf partners may be included as community representatives or informants or to co-lead working packages on co-creation, but may have limited opportunities to participate in leadership, funding control, and research agenda-setting. This imbalance can result in unequal recognition and advancement, even when all partners are committed to inclusive collaboration. While some hearing researchers may gain visibility and career opportunities through such projects, some deaf collaborators may find their contributions less formally acknowledged or supported.

As the field expands, simply involving deaf collaborators in more projects risks further concentrating burdens on a small pool of experts, intensifying co-creation fatigue. To avoid co-creation becoming a mere box-ticking exercise, projects must move beyond token representation and fundamentally reconsider how power, credit, and leadership are distributed. This means ensuring that deaf researchers, professionals, industry leaders, and organizations are embedded at every level of project design and implementation, with clear pathways for recognition, funding, and leadership development.

## 7 CONCLUSION: CO-CREATION AS A TRANSFORMATIVE ACTIVITY

This article has provided an in-depth, situated analysis of co-creation dynamics within two large multinational consortia. While the size and scope of these projects make them significant cases for examining structural patterns in sign language AI research in Europe, we acknowledge that our findings are specific to these two projects. At the same time, we cautiously situate these findings within broader conversations on co-creation in accessibility and sign language AI research, recognizing both the particularities of our cases and their relevance to ongoing challenges in the field.

Co-creation in sign language technology development presents numerous challenges, but the involvement of deaf-led organizations has been instrumental in making these projects more inclusive and representative. By addressing the inconvenient truths of co-creation, future projects can build on these lessons to ensure that co-creation is meaningful: recognize and resource the invisible labor of deaf partners, manage expectations through transparent and accessible communication, crip co-creation by dismantling structural ableism, diversify methods to address co-creation fatigue and intersectionality, and shift to deaf leadership and structural redistribution.

Our findings suggest that the formation of consortia is a critical juncture for embedding co-creation principles. Ensuring deaf leadership from the outset is not a matter of individual goodwill but of addressing structural conditions through which projects are conceived, funded, and governed – conditions that too often reproduce existing power asymmetries. This implies a reorientation of the role of hearing researchers: rather than leading consortia that lack meaningful deaf leadership, they can actively work to establish such leadership as foundational to project design. While individual awareness remains important – the EUD, for instance, has observed a shift in several hearing researchers' attitudes since the end of both projects – systemic change depends equally on external pressures. Horizon 2020's requirement for end-user involvement was a decisive factor in the EUD's participation in these projects, illustrating how top-down funding mandates and bottom-up researcher commitment must operate in tandem to achieve sustainable co-creation.

Yet co-creation is not only about improving project structures. It should be understood as an inherently transformative activity – one that actively levels the playing field and disrupts the status quo. This requires, among other things, creating pathways for training opportunities for deaf people in fields such as AI and machine learning, Human-Computer Interaction, linguistics, and Deaf

Studies, as an integral component of co-creation processes. Increasing the number of deaf researchers in these fields should be considered a necessary outcome of co-creation for any sign language technology development project. Additionally, increasing AI literacy among individual deaf signers and National Associations of the Deaf (NADs), which are frequently asked to participate in these projects, is essential. NADs must be equipped to engage with developments in SLMT in their signed language(s), gain a foundational understanding of these advancements, and identify knowledge gaps. This will enable them to provide informed responses on use cases, ethical considerations, and quality standards [17]. Beyond organizations, individual deaf end-users also need access to knowledge that allows them to critically engage with AI-driven technologies, recognizing both their affordances and limitations. Without these critical, transformative actions, co-creation risks becoming merely performative – a superficial gesture rather than a meaningful shift towards more equitable and inclusive research practices.